\documentclass[review,12pt]{elsarticle}
\usepackage{lineno,hyperref}
\usepackage{array}
\usepackage{amssymb}
\usepackage{enumerate}
\usepackage{makecell}
\usepackage{mathtools}
\usepackage{array}
\usepackage{multirow}
\usepackage{ulem}
\usepackage{color}
\usepackage{diagbox}
\usepackage{enumerate}
\usepackage{graphicx}
\usepackage{subfigure}
 \usepackage{tikz,mathpazo} 
\usepackage{indentfirst}
\usepackage{soul}
\usepackage{amsmath}
\usepackage[justification=centering]{caption}
\usepackage{tabularx}
\usepackage{algorithm}
\usepackage{algorithmicx}
\usepackage{algpseudocode}
\usepackage{adjustbox}
\usepackage{geometry}
\usetikzlibrary{shapes.geometric, arrows}
\modulolinenumbers[5]

\soulregister{\cite}7
\soulregister{\ref}7

\newcommand{\circledsmall}[1]{\lower.7ex\hbox{\tikz\draw (0pt, 0pt)%
    circle (.5em) node {\makebox[0.1em][c]{\small#1}};}}

\newcommand{\circledtiny}[1]{\lower.7ex\hbox{\tikz\draw (0pt, 0pt)%
    circle (.3em) node {\makebox[0.1em][c]{\tiny #1}};}}

\journal{Journal of \LaTeX\ Templates}

\begin{document}

\begin{frontmatter}

\title{Residual Feature-Reutilization Inception Network for Image Classification}

    \author[label1,label2]{Yuanpeng He\corref{cor1}}
    \author[label3]{Wenjie Song}
    \author[label5]{Lijian Li}
    \author[label4]{Tianxiang Zhan}
    \author[label1,label2]{Wenpin Jiao}
    
    \affiliation[label1]{organization={Key Laboratory of High Confidence Software Technologies (Peking University), Ministry of Education},
    	city={Beijing},
    	postcode={100871},
    	country={China}}
    
    \affiliation[label2]{organization={School of Computer Science, Peking University},
    	city={Beijing},
    	postcode={100871},
    	country={China}}
    
    \affiliation[label3]{organization={Nanhu Laboratory},
    	city={Jiaxing},
    	postcode={314000},
    	country={China}}

    \affiliation[label4]{organization={School of Computer and Information
    		Science, Southwest University},
    	city={Chongqing},
    	postcode={400715},
    	country={China}}

    \affiliation[label5]{organization={Department of Computer and Information Science, University of Macau},
    	city={Macau},
    	postcode={999078},
    	country={China}}

    \cortext[cor1]{Corresponding author: Yuanpeng He is with Key Laboratory of High Confidence Software Technologies (Peking University), Ministry of Education, Beijing, China; School of Computer Science, Peking University, Beijing, China
. E-mail address: heyuanpengpku@gmail.com and heyuanpeng@stu.pku.edu}

\begin{abstract}
Capturing feature information effectively is of great importance in the field of computer vision. With the development of convolutional neural networks (CNNs), concepts like residual connection and multiple scales promote continual performance gains in diverse deep learning vision tasks. In this paper, we propose a novel CNN architecture that it consists of residual feature-reutilization inceptions (ResFRI) or split-residual feature-reutilization inceptions (Split-ResFRI). And it is composed of four convolutional combinations of different structures connected by specially designed information interaction passages, which are utilized to extract multi-scale feature information and effectively increase the receptive field of the model. Moreover, according to the network structure designed above, Split-ResFRI can adjust the segmentation ratio of the input information, thereby reducing the number of parameters and guaranteeing the model performance. Specifically, in experiments based on popular vision datasets, such as CIFAR10 ($97.94$\%), CIFAR100 ($85.91$\%) and Tiny Imagenet ($70.54$\%), we obtain state-of-the-art results compared with other modern models under the premise that the model size is approximate and no additional data is used. 
\end{abstract}


\end{frontmatter}

\section{Introduction}
In recent years, we've witnessed a rapid advance of computer vision which is of great significance to aspects of human life. Generally, deep learning has contributed to this field a lot. The most representative deep neural network architectures in computer vision can be roughly divided into transformer-based and CNN-based models. Transformer is originally proposed for natural language processing, which has been transferred to vision tasks and achieves considerably satisfying performance recently. Specifically, vision transformer \cite{DBLP:journals/pami/00020C0GLTXXXYZ23} first introduces attention mechanism into computer vision whose strategy of information interaction enlargers the effective receptive field of related models observably so that crucial information can be better obtained. Due to efficiency of this architecture, the variations of transformer are devised corresponding to specific demands, and there are two main categories in the thoughts about improvements on the variations, namely integration of transformer framework with other models which are for particular usages and modifications on the original architecture. With respect to the former, DS-TransUNet \cite{DBLP:journals/tim/LinCXZLZ22} is a typical example, which synthesizes dual transformer-based architectures and U-Net to realize a breakthrough in medical image segmentation. Besides, some works focus on improvements on architecture of transformer, for instance, Mix-ViT \cite{DBLP:journals/pr/YuWZG23} tries to design a mix attention mechanism to create more sufficient passages for information interaction.

\begin{figure}[htbp]
	\centering
	\includegraphics[width=0.35\linewidth]{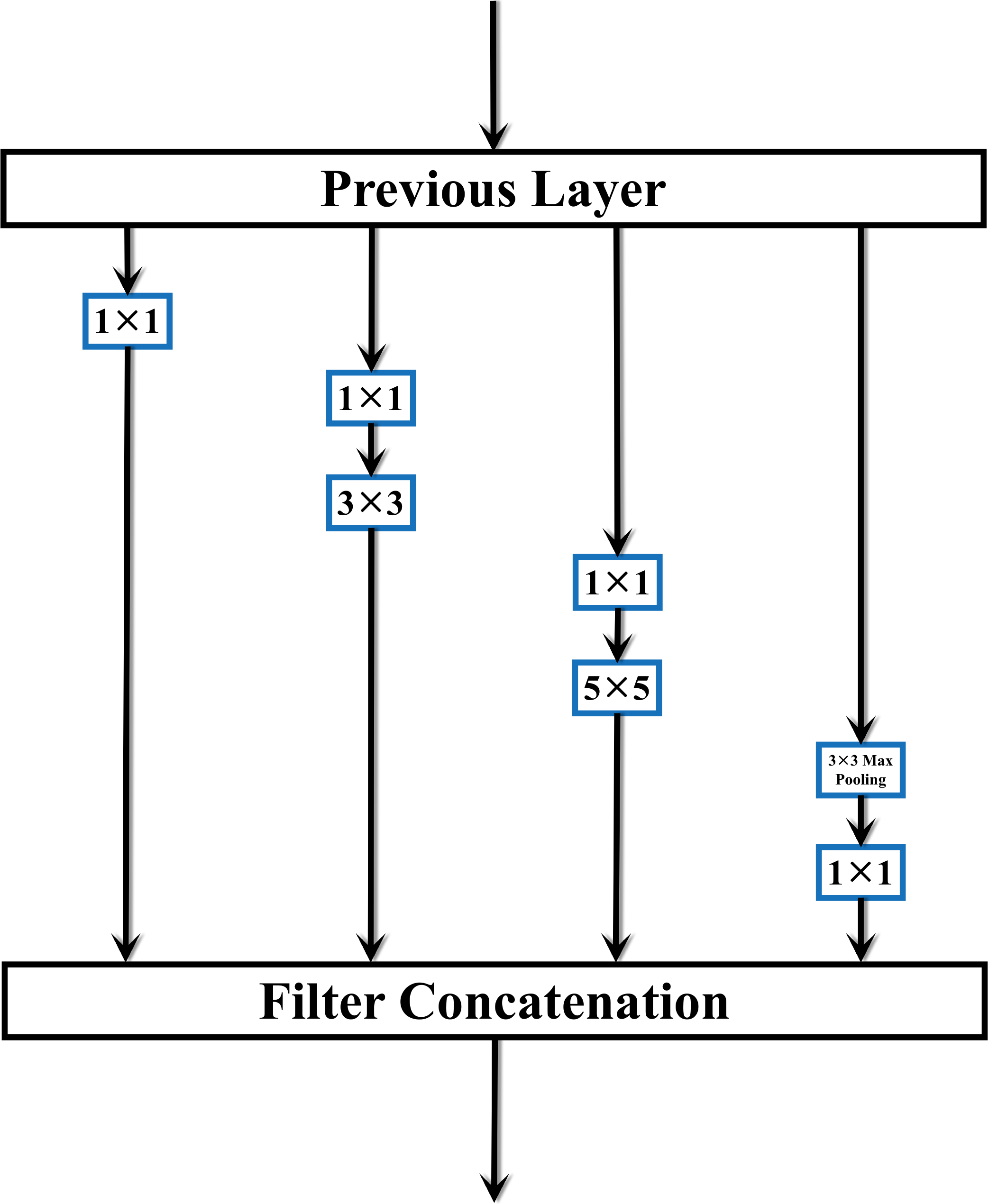}
	\caption{Original Inception from GoogLeNet}
	\label{Structure of Original Inception}%?????????
\end{figure}
Besides, the convolutional neural network has a longer history than transformer and is still favored by many researchers \cite{DBLP:journals/tip/DongLDZ22}. It is noteworthy that a performance milestone of CNN is the proposal of residual network \cite{DBLP:conf/cvpr/HeZRS16}. The invention of residual network makes training deeper structured CNNs possible. Nowadays, lots of vision models still benefit from this design to satisfy different requirements from various fields such as medical diagnosis \cite{DBLP:journals/pr/DentamaroGIMP22}. Specifically, some classical algorithms also get a second life with the addition of residual connection. For example, U-Net with nested residual connections realizes excellent performance on fault detection \cite{DBLP:journals/tgrs/GaoHZ22}. Although CNN has made significant progress, the performance of these networks is still restricted due to relatively small receptive field. In order to make up for this shortcoming, some researchers choose to combine characteristics of CNN and architecture of transformer such as Convformer-NSE \cite{articleSongyu} which utilizes the information fusion mechanism of transformer to provide CNN with a larger receptive field. Meanwhile, how to expand the perception area of CNN without using other kinds of network structures has become a research focus due to popularity of transformer-based architecture. Remarkably, acquirement of features from multiple scales to obtain more information has been a feasible solution to enlarge receptive field of CNNs, which enables model to process features at separate levels and boosts performances of them therefore. The concept of multi-scale has already been introduced into separate vision tasks \cite{DBLP:conf/cvpr/QiKG0WCLJ21} and its superiority is fully demonstrated by various effective models. Among them, InceptionNets \cite{DBLP:conf/aaai/SzegedyIVA17} enlargers the receptive field by stacking parallel convolutions of different kernel sizes. Besides, on the basis of common residual block \cite{DBLP:conf/cvpr/HeZRS16}, a multi-scale architecture called Res2Net \cite{DBLP:journals/pami/GaoCZZYT21} is devised to better obtain and aggregate information at different scales which replaces original $3\times3$ convolutions with a set of smaller convolution groups and proves traditional residual networks can also benefit from multi-scale design. The consideration of it resembles pyramid network \cite{DBLP:conf/cvpr/LinDGHHB17} which acquires multi-scale features through combination of high-level and low-level information. And the Res2Net block can continually enlarge the receptive field through stacking $3$$\times$$3$ convolutional layers in different groups of convolution to achieve satisfying performance as well. Advances in skeleton structure of modern visual models indicate that neural networks are tending to be more efficient and effective in multi-scale representation.

Obviously, obtaining information more effectively from a larger field of reception means a lot. Firstly, feature information is essential for image recognition and classification. In computer vision tasks, images often contain a large amount of pixels and color information, which is redundant for deep learning algorithms. Capturing effective feature information can help reduce the computational burden and storage space requirements, while improving the accuracy of classification and recognition. Secondly, the feature information can help computer to better understand image content. In computer vision, understanding semantic relations usually means recognizing objects, scenes, actions, etc. in images. By capturing the feature information, the computer can better comprehend the important information in images, which leads to more accurate classification and recognition. Third, feature information can enhance the generalization ability of vision model. In deep learning, generalization ability refers to the ability of a model to adapt to new data. By capturing effective feature information, model can better generalize to new image data improving the performance and accuracy on different tasks. Fourth, feature information can help solve the problem of data imbalance. In the field of computer vision, the data is often imbalanced, with some classes having a much larger number of samples than others. By capturing effective feature information, it can better distinguish samples from different categories, thereby addressing the problem of data imbalance. Besides, under the premise of a large amount of effective information, the use of better optimization algorithms can also improve the performance of vision models to some extent \cite{ghasemi2023geyser,ezugwu2022prairie,agushaka2023gazelle,hu2023detdo}. In conclusion, effectively capturing feature information is very important in computer vision to improve the accuracy of image recognition and classification, enhance the generalization ability of model, and solve the problem of data imbalance. 

\begin{figure*}[htbp]
	\begin{minipage}{0.49\linewidth}
		\centering
		\includegraphics[width=0.9\linewidth]{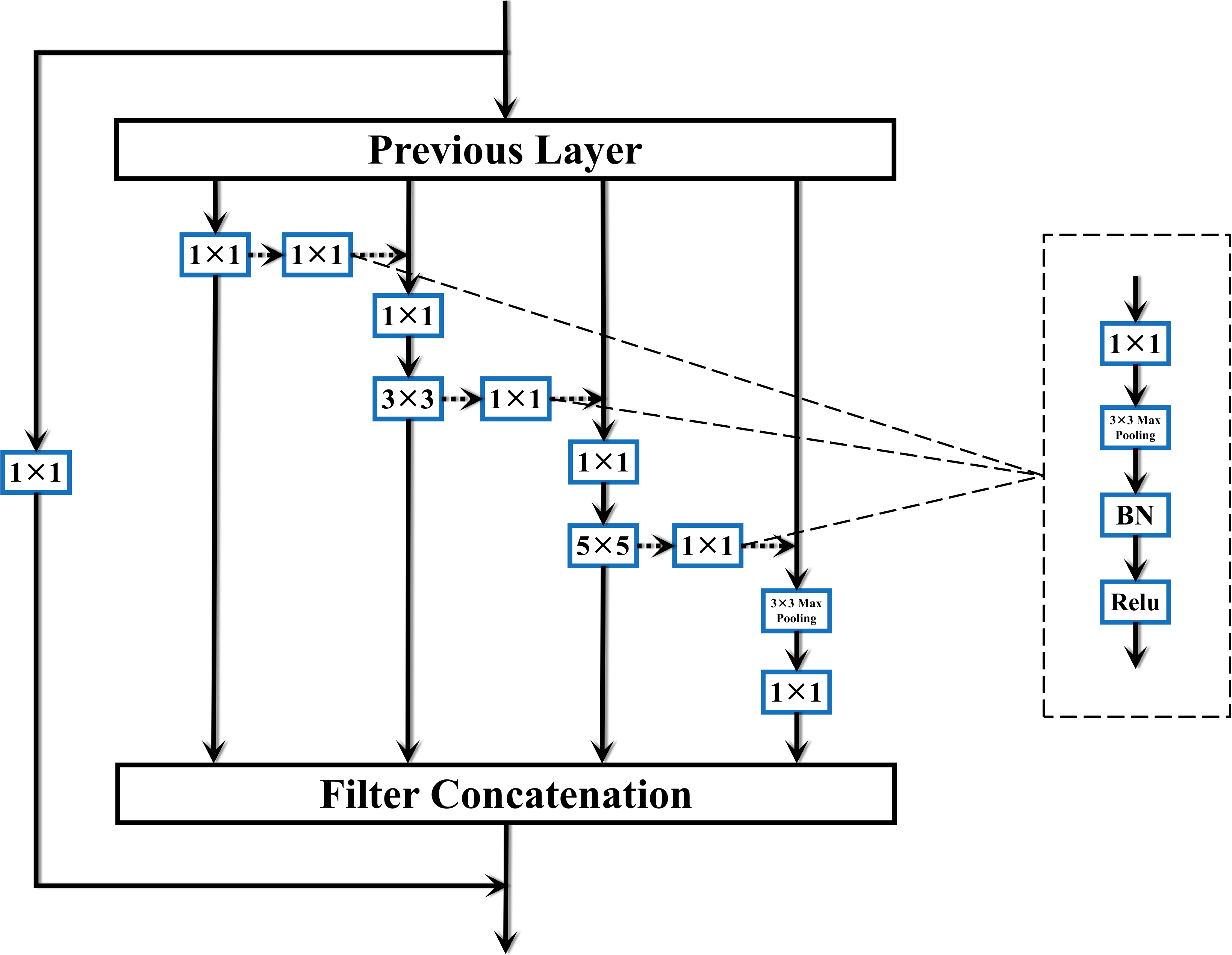}
		\caption{Residual Feature-Reutilization Inception}
		\label{Structure of Residual Feature-Reutilization Inception}%?????????
	\end{minipage}
	\begin{minipage}{0.49\linewidth}
		\centering
		\includegraphics[width=0.9\linewidth]{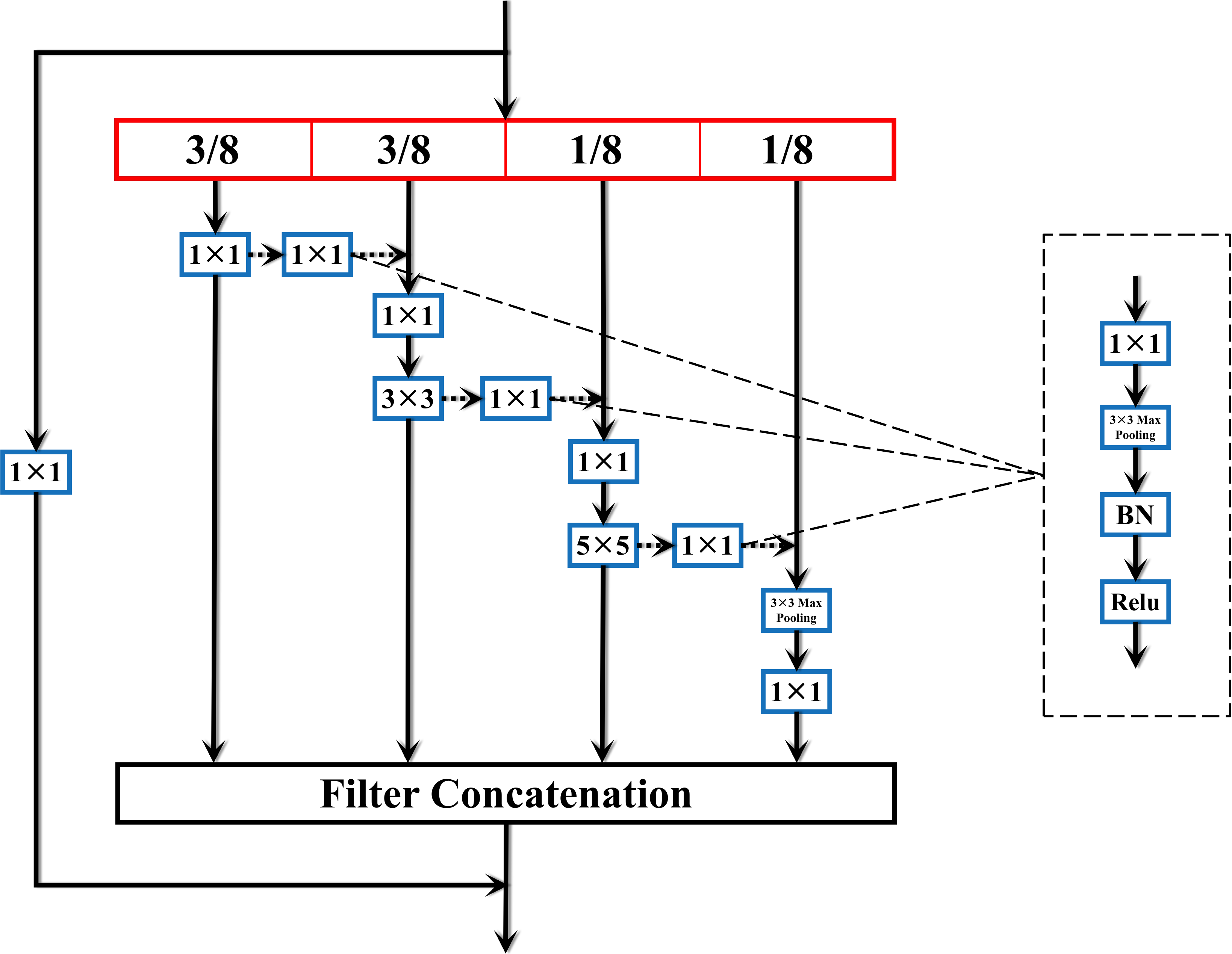}
		\caption{Split-Residual Feature-Reutilization Inception}
		\label{Split structure of Residual Feature-Reutilization Inception}%?????????
	\end{minipage}
	\label{comparison}
\end{figure*}
In this paper, we propose a high-efficiency and comprehensible multi-scale CNN model. Unlike previous models that enhance the ability of perception by deepening the number of network layers or using multiple identical receptive fields at a finer granularity, we choose to stack convolutions with different kernel sizes on parallel paths to obtain feature information at separate scales, which is different from Res2Net and enables the model to capture crucial information more flexibly. The convolution groups in separate paths are connected in a pattern similar to residual connection, which can increase the number of different scales showing that the output features enable the latter convolution groups to obtain richer hierarchical information. Specifically, the proposed model has two versions which consists of residual feature-reutilization inception (ResFRI) and split-residual feature-reutilization inception (Split-ResFRI) respectively. For ResFRI block, we input the complete feature map into each group of convolutions; with respect to Split-ResFRI block, we split the input features into four different parts according to the numbers of channels designed in GoogLeNet \cite{DBLP:conf/cvpr/SzegedyLJSRAEVR15}. The operation of split will significantly reduce the number of parameters and decrease training time a lot, however, which will also lead to a slight accuracy loss under some circumstances. A group of convolution first extracts features from input and the processed information are sent to the next groups of convolution with corresponding input features. This procedure ends when all information is processed. And the residual connection-like passages between groups of convolution adopt $1$$\times$$1$ convolutional layers to sample and enable the model to obtain stronger non-linearity in the separate receptive fields avoiding increasing calculation complexity too much. In other words, the existence of this operation enables information to be reutilized and the changes on the structure provide multi-scale feature extraction and information fusion at different scales, which makes up for the problem that CNNs have relatively limited receptive fields. Besides, a residual connection is also devised to the proposed inception network to reduce difficulty of network optimization. Synthesizing the peculiarities mentioned before, the proposed network possesses relatively smaller model size and achieves higher performance simultaneously, which is experimentally verified by the results of image classification on popular vision datasets. In summary, the proposed model combines features of multiple models and possesses considerable advantages compared with other modern models. The details of inception of GoogLeNet, ResFRI and Split-ResFRI are provided in Fig.\ref{Structure of Original Inception} and Fig.\ref{Structure of Residual Feature-Reutilization Inception}, \ref{Split structure of Residual Feature-Reutilization Inception}. 

The main contribution of the ResFRI can be summed up in four points which are listed as below:
\begin{enumerate}
	\item A novel multi-scale CNN architecture ResFRI and Split-ResFRI are proposed to fully utilize features from different scales and enlarge the receptive fields with four specially designed separate convolutional structures and corresponding information interaction passages.
	\item Split-ResFRI reduces parameter amount and Flops with acceptable accuracy loss by dividing features into multiple groups with ratios referring to the setting of channel number of ResFRI.  
        \item ResFRI and Split-ResFRI investigate the effect of pruning and pruning ratio on the performance of this model, which references the idea provided by CondenseNet \cite{DBLP:conf/cvpr/HuangLMW18}.
	\item Extensive experiments are conducted on popular image classification benchmark datasets, whose results demonstrate that our proposed methods achieves start-of-the-art performance compared with previous models which possess approximate model size and do not utilize extra data for training.
\end{enumerate}

\section{Residual Feature-Reutilization Inception Network}
\subsection{Introduction of Structure of ResFRI and Split-ResFRI}
\begin{figure}[h]
	\centering
	\includegraphics[width = 0.6\textwidth]{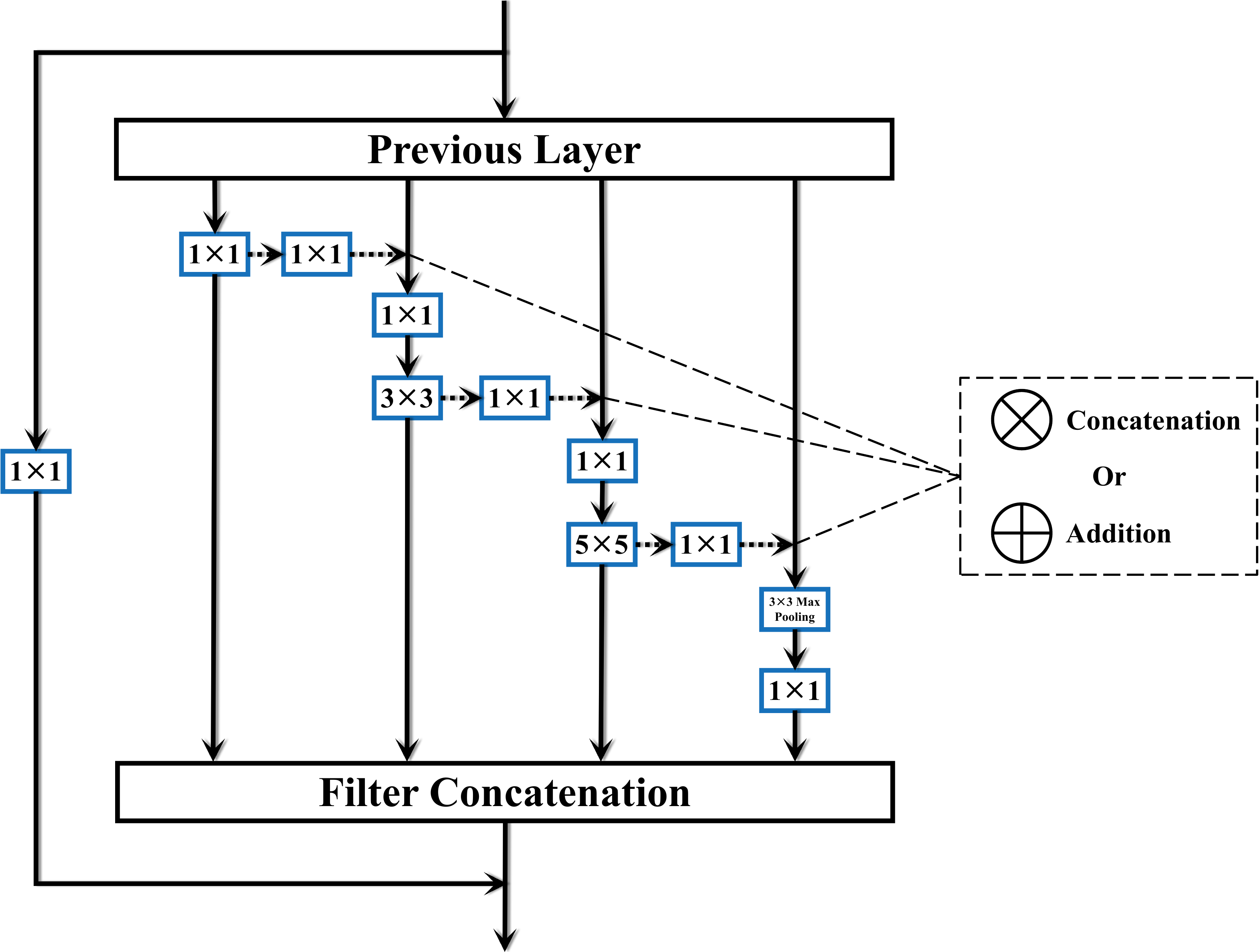}
	\caption{Details of Passages Between Convolution Groups.\\ \textbf{Note: The Split-ResFRI Also Adopts the Same Information Interaction Strategy as ResFRI}}
	\label{operation}
\end{figure}
The details of ResFRI and Split-ResPRI are presented in Fig.\ref{Structure of Residual Feature-Reutilization Inception} and \ref{Split structure of Residual Feature-Reutilization Inception}. Suppose information from previous layer as $\xi_{Pre}$ and the operations of convolutional layers as $Conv$, the main difference of ResFRI (RI) and Split-ResFRI (SRI) in processing of input can be defined as:
%\begin{equation}
%	\label{eq:alpha}
%	\hat{F}_{h}^{i}(x,y) =\left\{
%	\begin{array}{lcl}
	%		\quad 0,\quad   &if \quad |F_{h}^{i}(x,y)| < \sigma_{L} \\
	%		F_{h}^{i}(x,y),\quad  &othersize 
	%	\end{array},(1 \leq i\leq 3L)
%	\right.
%\end{equation}

\begin{equation}
	\label{eq:alpha}
	\left\{
	\begin{array}{lcl}
		Conv(\xi_{Pre},\xi_{Pre},\xi_{Pre},\xi_{Pre}), &RI \\\\
		Conv(\gamma_{1},\gamma_{2},\gamma_{3},\gamma_{4})\\ \gamma_{1,2} = 3*\xi_{Pre}//8 \quad \gamma_{3,4} = \xi_{Pre}//8, &SRI
	\end{array}
	\right.
\end{equation}

where ResFRI directly processes information contained in all channels from last layer with four different convolutional groups and Split-ResFRI receives information from 3/8, 3/8, 1/8 and 1/8 channels, which is devised according to original settings of GoogLeNet \cite{DBLP:conf/cvpr/SzegedyLJSRAEVR15}. Compared with the primitive structure of inception contained in GoogLeNet, some improvements are dexterously designed in ResFRI and Split-ResFRI. To be specific, in order to reuse information, we construct passages between adjacent groups of convolutional layers. Moreover, a residual connection is also devised to reduce difficulty of network optimization and to avoid problems like overfitting and abnormal gradients. Besides, to match feature channels between groups of convolutional layers and residual connection to final output, a structure consists of $1$$\times$$1$ Convolutional layers, $3$$\times$$3$ MaxPool, BatchNorm and ReLu (cmbr) is utilized. It also further enhances extraction of information by realizing cross channel information combination and non-linear feature transference. And it's worth noting that the information combination is achieved by adding or concatenating features and the operation is described in Fig.\ref{operation}. Suppose the information processed by former group of convolutional layer as $\delta$ and the input to this group as $\kappa$, then the fusion of information between groups of convolutional layer can be defined as:
\begin{equation}
	\label{eq:alpha}
	\mathbb{F} =\left\{
	\begin{array}{lcl}
		Addition(cmbr(\delta), \kappa) \\
		Concat(cmbr(\delta), \kappa)
	\end{array}
	\right.
\end{equation}
where $Addition$ represents addition of $cmbr(\delta)$ and $\kappa$, and $Concat$ denotes concatenation of them. Moreover, the comparison of performance and resource consumption between these methods can be found in the ablation study based on ResFRI.

To reduce consumption of computation resources, we discard the $3$$\times$$3$ convolutional layers designed by Res2Net and comply with the original design of inception of GoogLeNet. Besides, we notice that the idea of connections between different groups of convolutional layers is very similar to the one of DenseNet \cite{DBLP:journals/pami/HuangLPMW22}, the extra passages may help improve performance of network. However, CondenseNet\cite{DBLP:conf/cvpr/HuangLMW18} points out that the dense connections are probably redundant under certain circumstances and this phenomenon may reduce accuracy and efficiency of network. As a result, we prune newly-constructed passages of information transference except the residual connection in ResFRI to avoid unnecessary calculations and obtain higher accuracy. More specifically, we adopt unstructured pruning which trims the single weight and does not require a whole row of pruning. The advantage is that the original accuracy can be maintained, because structured pruning is easy to cut out those important weights. Besides, the tools of pruning is provided by PyTorch which will abandon a part of weight parameters using mask matrices without changing the original size of models. In the last, for the filter concatenation ($\mathbb{FC}$) and synthesizing the operations defined above, suppose $Conv$ consists of $[\mathbb{C}_1,\mathbb{C}_2,\mathbb{C}_3,\mathbb{C}_4]$, it can be defined as:
\begin{equation}
	\label{eq:alpha}
	\mathbb{FC} =\left\{
	\begin{array}{lcl}
		Concat(\mathbb{C}_1(\xi_{Pre}),\mathbb{C}_2(\mathbb{F}(\mathbb{C}_1(\xi_{Pre})),\xi_{Pre}),\\ \mathbb{C}_3(\mathbb{F}(\mathbb{C}_2(\mathbb{F}(\mathbb{C}_1(\xi_{Pre})),\xi_{Pre})),\xi_{Pre}),\\ \mathbb{C}_4(\mathbb{F}(\mathbb{C}_3(\mathbb{F}(\mathbb{C}_2(\mathbb{F}(\mathbb{C}_1(\xi_{Pre})),\xi_{Pre})),\xi_{Pre})),\xi_{Pre})), \ RI\\\\
		Concat(\mathbb{C}_1(\gamma_1),\mathbb{C}_2(\mathbb{F}(\mathbb{C}_1(\gamma_1)),\gamma_2),\\ \mathbb{C}_3(\mathbb{F}(\mathbb{C}_2(\mathbb{F}(\mathbb{C}_1(\gamma_{1})),\gamma_{2})),\gamma_{3}),\\ \mathbb{C}_4(\mathbb{F}(\mathbb{C}_3(\mathbb{F}(\mathbb{C}_2(\mathbb{F}(\mathbb{C}_1(\gamma_{1})),\gamma_{2})),\gamma_{3})),\gamma_{4})), \ SRI
	\end{array}
	\right.
\end{equation}

\subsection{Discussions on ResFRI and Split-ResFRI and multi-scale feature fusion strategy}
The proposed CNN architecture incorporates two key components: Residual Feature Reutilization Inceptions (ResFRI) and Split-Residual Feature-Reutilization Inceptions (Split-ResFRI). These components are designed to enhance information flow and passages between convolutional layers, enabling the network to extract richer feature representations and optimize overall network performance.

ResFRI is a modified version of the popular ResNet and GoogleNet. In ResFRI, the input feature map is fed into two paths respectively: a main path and a residual path. In detail, the main path performs convolutional combinations of different structures for multi-scale information acquisition with mutual interaction, while the residual path preserves the original feature map and reutilizes it as residual features. These residual features are then added back to the main path's output, enabling the network to realize better optimization performance and improve the understanding of information processed. By stacking the ResFRI module, the whole network is capable of obtaining information at more flexible level, which allows to capture more comprehensive feature representations and better preserve fine-grained details.

Besides, Split-ResFRI extends the concept of ResFRI by further splitting feature map according to the channel setting of ResFRI. Different from ResFRI, each group of convolutions only receive a part of information for reduction of network parameters and demand of computation, while retaining the ability of extraction of features, which allows the network to optimize information flow and capture a broader range of feature representations for improved performance. To be specific, same as the design of ResFRI, there exist passages between adjacent groups of convolution which are devised for feature reutilization. And the passages are composed of a $1$$\times$$1$ convolution for matching the number of channels of the convolutional group that will be input, $3$$\times$$3$ maxpooling for realizing further dimensionality reduction and removal of redundant information, BatchNorm layer and Relu activation function for solving the gradient problem and improve the training speed and stability. And the final information fusion are realized by adding or concatenating feature maps in passages and groups of convolutions. 

Due to the completely different convolution operations in various convolutional combinations, they have different receptive fields that enable them to extract different granularity feature information which provides different context information at different scales. When the information in different convolutional combinations is fused through the information interaction passages mentioned above, the feature of different granularities can be further combined to realize the reorganization of multi-scale information and integrate aforementioned context information to provide a more comprehensive and rich visual context. As a result, the network can extract richer semantic associations from the optimized information, and achieve a deeper mining of features with different granularities. In sum, the designed approach allows for more effective information flow, improving feature extraction capabilities, and better generalization performance for a wide range of image classification tasks.

\subsection{Other Important Settings of ResFRI and Split-ResFRI}
During the process of experiment, we notice that the MaxPool layers may hamper the network to capture information effectively and weaken performance of it, we argue that the MaxPool layers may destruct information contained in the relative low-resolution figures instead of being helpful in extraction of features. Verified by experiments, we change the MaxPool layer into AvgPool layer eventually.

Raised by \cite{DBLP:conf/cvpr/HuangLMW18}, the dense connections may have negative impact on the process of learning and decrease accuracy of models. Therefore, we try to cancel some passages between convolution groups to avoid too dense connections in the ResFRI and Split-ResFRI utilizing different pruning ratio. Finally, we set the drop rate of passages of information transference to $0.7$ and $0$ on addition and concatenation version of ResFRI respectively, which can be defined as:
\begin{equation}
	\label{eq:alpha}
	Pruning\ Ratio =\left\{
	\begin{array}{lcl}
		0.7,\quad   &Addition,RI \\
		0,\quad   &Concatenation,RI
	\end{array}
	\right.
\end{equation}

With respect to Split-ResFRI, because of underlying performance loss which may be brought by segmentation of information, we set the pruning rate uniformly to $0$ in order to strengthen information interaction among groups of convolutional layers. And it is worth noting that when the classes contained in datasets are becoming more, we are supposed to reduce the amount of pruning to better promote information transference for the version of addition of ResFRI, which can be illustrated in the following experiments on vision datasets. In the last, the results in the part of ablation study will prove the effectiveness of these modifications based on ResFRI.

\subsection{Limitations of existing methods}
ResNet mainly performs feature extraction by stacking deep networks, and does not explicitly introduce multi-scale information in the network, which largely limits its performance improvement. In order to solve the problem of multi-scale information acquisition, Res2Net introduces multiple branches to process information at different scales. It is successful to some extent, but the size of the convolution kernel in each branch is usually fixed, which means that in each branch, there is a limit on the size of the receptive field. For some specific scenarios, larger or smaller receptive fields may be required to effectively capture the features in vision information. But similar to ResNet, the residual connections it has do guarantee the performance of the model. For the inception series of networks, researchers have made various improvements. For example, a design similar to residual connection is introduced into the inception networks, or more different branches are utilized to obtain different levels of feature information, so that the network can understand the purpose of each visual task more efficiently. However, they often lack a similar information interaction design as Res2Net which simply concatenates the information of each branch and then fuses it through a convolutional layer, without considering the potential importance of information interaction between different branches. This is a key factor where the promotion of their model performance is limited. And, it is worth pointing out that due to the depth and complexity of the above models, the training and inference process requires a lot of computational resources and storage space. Deeper ResNet models may take longer time to train and have a corresponding increase in GPU memory requirements. This makes it difficult to deploy and run these models in resource-limited environments, such as mobile devices or embedded systems, and the models are more space intensive in terms of storage and transmission. Therefore, in response to address these problems, we propose a convolutional neural network structure that can balance model size, training time and performance, which is able to achieve very excellent accuracy on classification task under the premise of usage of very limited resources.

\section{EXPERIMENTS}
\begin{figure*}[htbp]
	\centering
	\includegraphics[width = 0.95\textwidth]{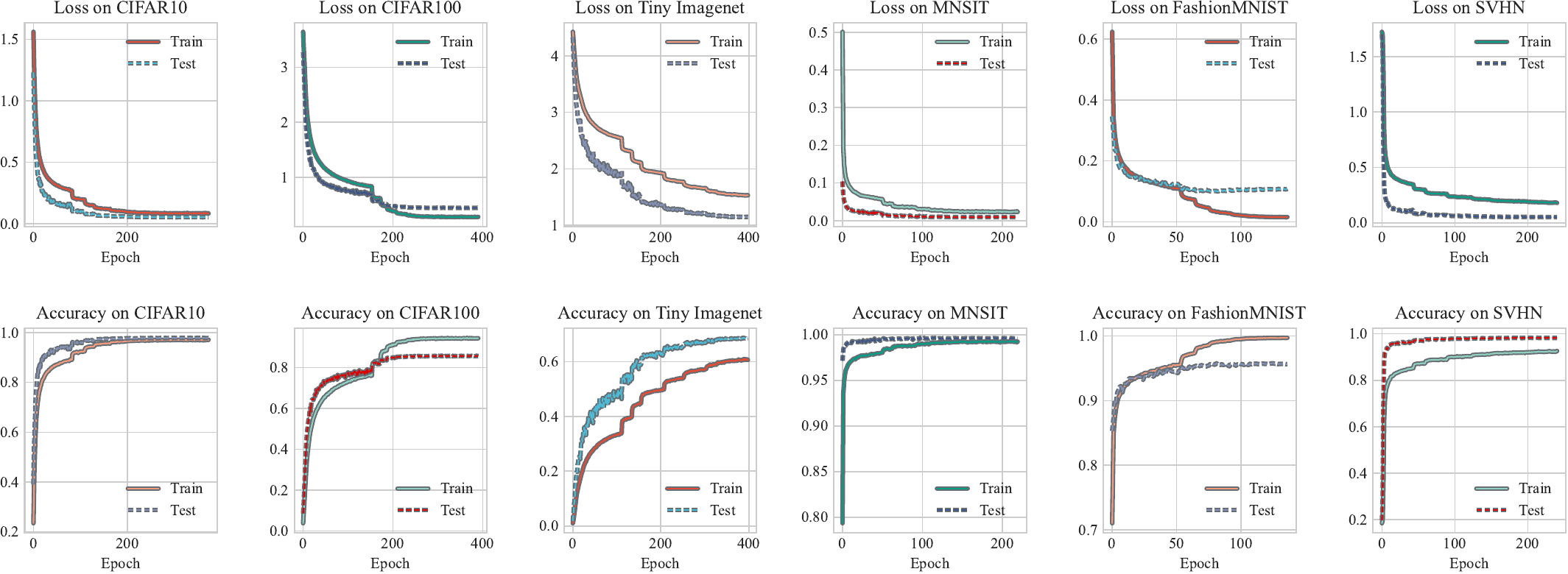}
	\caption{Loss and Accuracy of ResFRI-addition on Datasets}
	\label{tendency}
\end{figure*}
\subsection{Implementation Details}
We implement the whole framework of the proposed model utilizing code framework provided by PyTorch. And in order to ensure fairness of comparison among different methods, we provide experiment results of classical and newly proposed models with or without extra data. Due to our limited computation resources, apart from necessary ablation experiments, we choose the task of image classification on the popular datasets, such as CIFAR10, CIFAR100, Tiny Imagenet, MNIST, FashionMNIST and SVHN. And in the process of training on one RTX $4090$ and $3060$ GPUs, we use the optimizer SGD with momentum $0.9$, weight decay $0.0005$, batch size $64$. We adopt the same data augmentation strategy as \cite{DBLP:conf/cvpr/HeZRS16} and images are resized into $32\times32$ except for Tiny ImageNet dataset in which the model size will variate to some extent. Moreover, the initial learning rate is set to $0.01$ and it is reduced by half if validation loss does not decrease within $10$ epochs. And tendency of accuracy and loss in the training process of ResFRI is given in Fig.\ref{tendency}.

\subsection{Metric of Image Classification}
Image classification accuracy is the proportion of a model that predicts the class of a given image correctly. In computation, it is usually defined as the ratio of the number of all correctly classified images (i.e., true examples) to the total number of images. More academically, accuracy refers to the proportion of all samples in which the judgment result is consistent with the actual result (TP+TN). In the task of image classification, it reflects the model's confidence that the image is of a certain class. Specifically, the calculation formula can be defined as follows:
\begin{equation}
    Acc = \frac{TP+TN}{TP + FP + TN + FN}
\end{equation}

Correspondingly, the error rate can be defined as:
\begin{equation}
    Err = 1 - Acc
\end{equation}

Generally speaking, people use these two metrics to represent the performance of a model for image classification.
\subsection{Experiments on CIFAR-10}
\begin{figure*}[htbp]
	\centering
	\includegraphics[width = 0.95\textwidth]{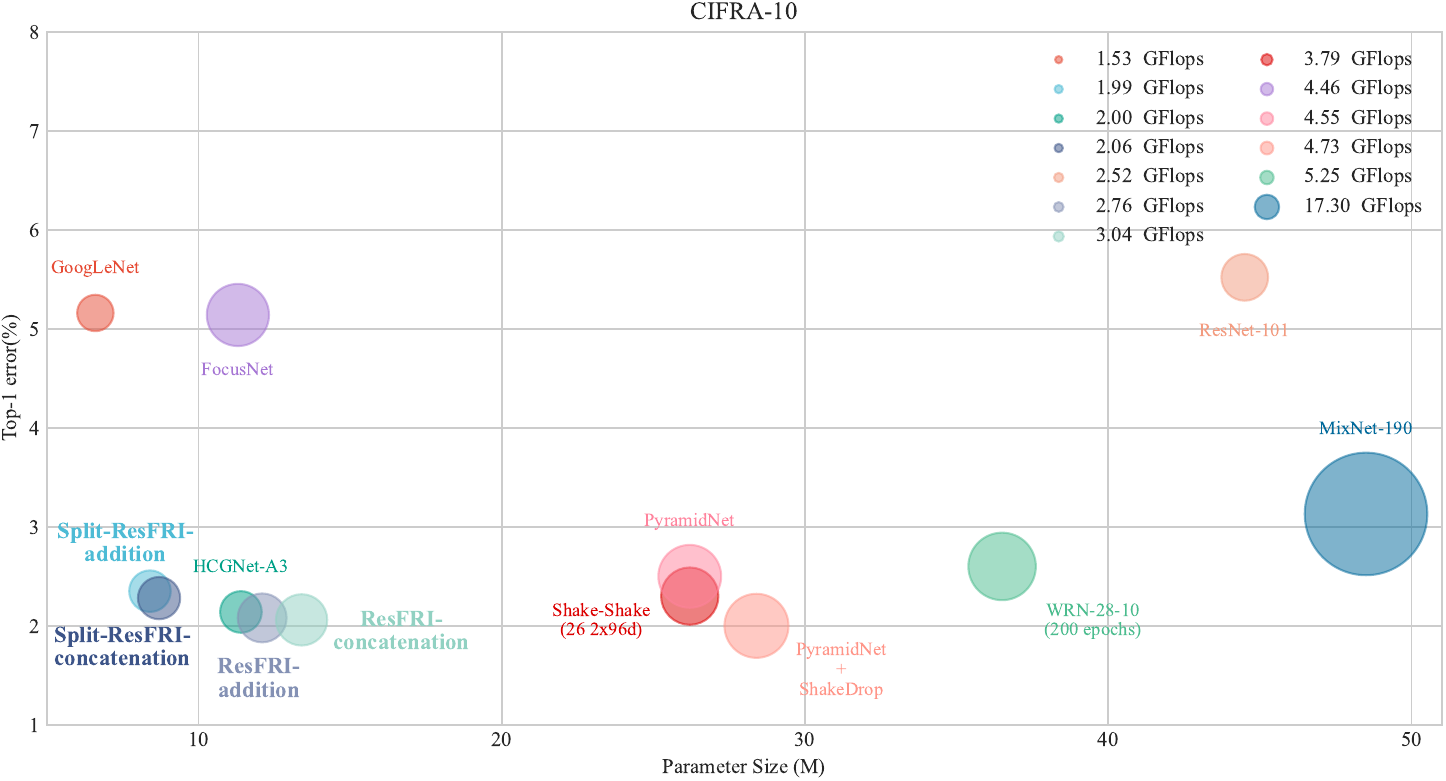}
	\caption{Comparisons of models on CIFAR10 Dataset }
	\label{CIFAR10_pop}
\end{figure*}

The CIFAR10 dataset contains $50$k training images and $10$k testing images from $10$ classes whose resolution is $32$$\times$$32$. And the detailed results of comparisons of different models will be clearly provided in Table \ref{table1} and Fig.\ref{CIFAR10_pop}.

\begin{table*}[htbp]\scriptsize
	\centering
	\caption{Error rate (\%) and Model Size on the CIFAR-10 Dataset}
	\renewcommand\arraystretch{1.4} %??????
	\begin{tabular}{cccccccc} %????????c
		\hline 
		\multicolumn{4}{l}{Method} & Flops&  Params  &    top-1 err. & Accuracy \\ \cline{1-8}
		\multicolumn{4}{l}{ResNet-101 \cite{DBLP:conf/cvpr/HeZRS16}}&2.52G &     44.5M    &      5.52   & 94.48     \\
		\multicolumn{4}{l}{GoogLeNet \cite{DBLP:conf/cvpr/SzegedyLJSRAEVR15}}&1.53G &     6.6M    &      5.16 & 94.84       \\
		%\multicolumn{4}{l}{ResNeXt-29, 16$\times$32d \cite{DBLP:conf/cvpr/XieGDTH17}}&4.05G&     25.2M    &      3.87       \\
		%\multicolumn{4}{l}{ResNeXt-29, 8$\times$64d \cite{DBLP:conf/cvpr/XieGDTH17}}&5.41G&     34.4M    &      3.65        \\
		%\multicolumn{4}{l}{ResNeXt-29, 16$\times$64d \cite{DBLP:conf/cvpr/XieGDTH17}}&10.73G &     68.1M    &      3.58       \\
		%\multicolumn{4}{l}{	CapsNet \cite{DBLP:conf/nips/SabourFH17}}&- &     -   &      10.6    \\
		%\multicolumn{4}{l}{	DropConnect \cite{DBLP:conf/icml/WanZZLF13}}& -&     -   &      9.32      \\
		\multicolumn{4}{l}{	RMDL \cite{DBLP:conf/icisdm/KowsariHBMB18}}&- &     -   &      8.74 & 91.26     \\
		%\multicolumn{4}{l}{	FractalNet \cite{DBLP:conf/iclr/LarssonMS17}}&- &     38.6M   &      7.27      \\
		\multicolumn{4}{l}{	SOPCNN \cite{DBLP:conf/mldm/Assiri19}}&- &     4.2M   &      5.71 & 94.29      \\
		\multicolumn{4}{l}{DenseNet-BC (k=24) \cite{DBLP:journals/pami/HuangLPMW22}}&- &    15.3M     &      5.19   & 94.81     \\
		\multicolumn{4}{l}{FocusNet \cite{DBLP:journals/pr/ZhangSS22}}&4.46G&    11.3M     &      5.14  & 94.86     \\
		\multicolumn{4}{l}{LOW \cite{DBLP:journals/pr/SantiagoBSCN21}}&-&    -     &      4.8  & 95.2    \\
		\multicolumn{4}{l}{DPN-28-10 \cite{DBLP:conf/aaai/YangAZHZXLX20}}&- &     47.8M    &      3.65  & 96.35      \\
		\multicolumn{4}{l}{DCDN \cite{DBLP:journals/prl/PatelW22}}&- &     -   &       3.54 & 96.46 \\
		\multicolumn{4}{l}{NASNet-A \cite{DBLP:conf/aaai/YangAZHZXLX20}}&- &      3.3M    &       3.41 & 96.59      \\
		\multicolumn{4}{l}{AmoebaNet-A \cite{DBLP:conf/aaai/YangAZHZXLX20}}&- &      4.6M    &       3.34       & 96.66\\
		\multicolumn{4}{l}{AOGNet \cite{DBLP:conf/aaai/YangAZHZXLX20}}&- &     24.8M    &      3.27  & 96.73    \\
		\multicolumn{4}{l}{MixNet-190 \cite{DBLP:conf/aaai/YangAZHZXLX20}}&17.3G &     48.5M    &      3.13     & 96.87 \\
		\multicolumn{4}{l}{AmoebaNet-B \cite{DBLP:conf/aaai/YangAZHZXLX20}}&- &      34.9M    &       2.98      & 97.02 \\
		%\multicolumn{4}{l}{OR-WideResNet \cite{DBLP:conf/cvpr/ZhouYQJ17}}&- &     18.2M    &       2.98       \\
		%		\multicolumn{4}{l}{WRN-28-10 (200 epochs)}&5.25 GFlops &    36.5M    &       2.6      \\
		\multicolumn{4}{l}{NAONet \cite{DBLP:journals/pr/HuBAYF22}}&- &    -  &     2.65 & 97.35  \\
		\multicolumn{4}{l}{WRN-28-10  (200 epochs) \cite{DBLP:conf/icml/KwonKPC21}}&5.25G &    36.5M    &       2.6  & 97.4    \\
		\multicolumn{4}{l}{PyramidNet \cite{DBLP:conf/cvpr/LinDGHHB17}}&4.55G &    26.2M    &       2.5  & 97.5    \\
		\multicolumn{4}{l}{Shake-Shake (26 2$\times$96d) \cite{DBLP:conf/iclr/ForetKMN21}}&3.79G &    26.2M    &       2.3 & 97.7     \\
		\multicolumn{4}{l}{HCGNet-A3 \cite{DBLP:conf/aaai/YangAZHZXLX20}}& 2.0G&11.4M&2.14 & 97.86\\
				%\multicolumn{4}{l}{HCGNet-A3 \cite{DBLP:conf/aaai/YangAZHZXLX20}}&2.0 GFlops &     11.4M   &      2.14       \\
		\multicolumn{4}{l}{PyramidNet+ShakeDrop \cite{DBLP:conf/iclr/ForetKMN21}}&4.73G &     28.4M   &      2.1 & 97.9      \\\cline{1-8}
		%		\multicolumn{4}{c}{Shake-Shake (26 2x96d)} &     -    &     14.30       &-\\\cline{1-7}
		\multicolumn{4}{l}{ResFRI-addition}& 2.76G &      12.1M   &   2.08 & 97.92        \\
		\multicolumn{4}{l}{Split-ResFRI-addition}& 1.99G &      8.4M   &      2.35 & 97.65       \\
		\multicolumn{4}{l}{ResFRI-concatenation}&3.04G &     13.4M     &  \textbf{2.06} & \textbf{97.94} \\
		\multicolumn{4}{l}{Split-ResFRI-concatenation}&2.06G &     8.7M     &   2.28 & 97.72\\
		\hline
	\end{tabular}
	\label{table1}    
\end{table*}

It can be obtained that the ResFRI and Split-ResFRI achieve relatively satisfying performance on CIFAR-10 dataset-based image classification task. Compared with traditional models like ResNet-101 \cite{DBLP:conf/cvpr/HeZRS16}, ResFRI and Split-ResFRI have better performance with much lower parameter amount. Although ResFRI-addition has $0.24$ GFlops and ResFRI-concatenation has $0.52$ GFlops higher than ResNet-101, we have a remarkable $3.44$\% and $3.46$\% performance gain on top-1 err while parameter amounts reduce by $32.4$M and $31.1$M. For Split-ResFRI, the version of addition has $0.53$ GFlops and $36.1$M parameters lower than ResNet-101, but we get $3.17$\% performance improvement.  Besides, Split-ResFRI-concatenation also has $0.46$ GFlops  and $35.7$M parameters lower than ResNet-101 and $3.24$\% higher accuracy. In sum, both of the Split-ResFRIs have lower Flops and parameter amounts and achieve better results than ResNet-101. Compared with two versions of ResFRI, Split-ResFRIs sacrifice a little bit of precision in exchange for a considerable reduction on Flops and parameter amount. For ResNeXt-29, it outperforms ResNet-101 using larger model scales, but it still trails by at least $1.23$\% in comparison with ResFRI and Split-ResFRI. And with respect to GoogLeNet \cite{DBLP:conf/cvpr/SzegedyLJSRAEVR15}, no matter it is ResFRI or Split-ResFRI, we all have achieved performance leadership. Besides, it is worth noting that both versions of Split-ResFRIs have similar flops and parameter amounts to GoogLeNet, but still possess a performance lead of over $2.8$ percentage. 

Moreover, when encountering some relatively new models, ResFRI and Split-ResFRI still prove their superiority on classification task. For DenseNet \cite{DBLP:journals/pami/HuangLPMW22}, it possesses a similar model scale to ResFRI-concatenation, but it has a $3.13$\% performance disadvantage in the final result. And with respect to FocusNet \cite{DBLP:journals/pr/ZhangSS22}, although it has approximate parameter amount to ResFRI, it falls behind by $2.92$\% at least compared with ResFRI. And Split-ResFRI is able to possess a more obvious performance advantage. Compared with ResFRI and Split-ResFRI, its disadvantage is still significant with performance trailing by at least $0.63$\%. Then, for WRN-28-10 \cite{DBLP:conf/icml/KwonKPC21}, PyramidNet \cite{DBLP:conf/cvpr/LinDGHHB17} and Shake-Shake($26$ $2$$\times$$96d$) \cite{DBLP:conf/iclr/ForetKMN21}, all of them have higher flops and parameter amount than ResFRI and Split-ResFRI, but the proposed models achieve better accuracy except for Split-ResFRI-addition meanwhile. However, we want to point out that Split-ResFRI-addition has far less GFlops and parameter amount than the above models for comparison. Moreover, we notice that PyramidNet+ShakeDrop \cite{DBLP:conf/iclr/ForetKMN21} has a a very approximate performance ($-0.04$\%) to ResFRI-concatenation, which is a very competitive opponent. However, the cost of the combination of PyramidNet and ShakeDrop is $71.3$\% higher flops and $134.7$\% larger parameter amount than Res-FRI-concatenation. We think this comparison also illustrates the advantage of the proposed method when considering performance and differences on computing resources consumption of the proposed models. In summary, the experiment on CIFAR-10 dataset strongly proves the effectiveness and validity of the proposed model on image classification task, and Split-ResFRI still has enough competitiveness when taking the reduction on Flops and the number of parameters by a significant amount into consideration. 

\subsubsection{Comparison with models using extra techniques and data}
\begin{table*}[htbp]\scriptsize
	\centering
	\caption{Error rate (\%) and Model Size on the CIFAR-10 Dataset}
	\renewcommand\arraystretch{1.4} %??????
	\begin{tabular}{ccccccc} %????????c
		\hline 
		 Method &     top-1 err. & Accuracy & Method with extra data  &    top-1 err. & Accuracy\\ \hline
		 WRN-28-10 (200 epochs) + SAM & 2.7 & 97.3 &  ViT-H/14 + JFT & 0.50 & 99.5 \\
		 WRN-28-10 (1800 epochs) + SAM & 2.4 & 97.6 & ViT-L/16 + JFT & 0.58 & 99.42\\
		 Shake-Shake (26 2x96d) + SAM & 2.3   & 97.7 &  ViT-L/16 + I21k &  0.85 & 99.15 \\
		 PyramidNet + SAM & 2.7 & 97.3 & ResNet152x4 + BiT-L   & 0.63 & 99.37\\ 
		 PyramidNet + ShakeDrop + SAM & 2.1  & 97.9 &  $\mu$2Net + ViT L/16 & 0.51 & 99.49 \\
		 EffNet-L2 + SAM + Pre-Training$^{*}$ & 0.3 & 99.7 &DINOv2 + ViT-g/14 & 0.50 & 99.5\\ \hline
             The Proposed Method&top-1 err. &Accuracy&The Proposed Method & top-1 err.&Accuracy\\\hline
            ResFRI-addition & 2.08 & 97.92 & Split-ResFRI-addition & 2.35 & 97.65 \\
            ResFRI-concatenation & 2.06 & 97.94 & Split-ResFRI-concatenation & 2.28 & 97.72 \\
            ResFRI-addition + SAM & 1.72 & 98.28 & Split-ResFRI-addition + SAM &1.88 & 98.12 \\
            ResFRI-concatenation + SAM &\textbf{1.65} & \textbf{98.35} & Split-ResFRI-concatenation + SAM &1.79 & 98.21 \\\hline

			\end{tabular}
	\label{}    
\end{table*}

In comparison with models using additional training techniques and data, our model still achieves relatively satisfying good performance. Among them, compared with the models using SAM \cite{DBLP:conf/iclr/ForetKMN21}, our proposed method can realize better image classification accuracy with a smaller number of model parameters and size. With respect to EffNet-L2 + SAM + Pre-Training, our model size is much smaller than EffNet-L2, and no additional data such as JFT is used, so it has a certain disadvantage in classification performance. When compared with the model utilizing much more parameters and computation resources \cite{DBLP:conf/icml/LutatiW23,DBLP:journals/corr/abs-2304-07193,DBLP:journals/corr/abs-2205-12755}, our model clearly does not have any advantage in image classification performance. Moreover, these models are also pre-trained on large datasets, which makes them have better generalization ability and achieve reasonably good classification results on such small datasets. In such cases, the amount of computing resource they need is very large. However, it is worth pointing out that our model is able to achieve state-of-the-art image classification accuracy with the same order of magnitude of model size and without using additional training data.

\subsection{Experiments on CIFAR-100}
\begin{figure*}[htbp]
	\centering
	\includegraphics[width = 0.95\textwidth]{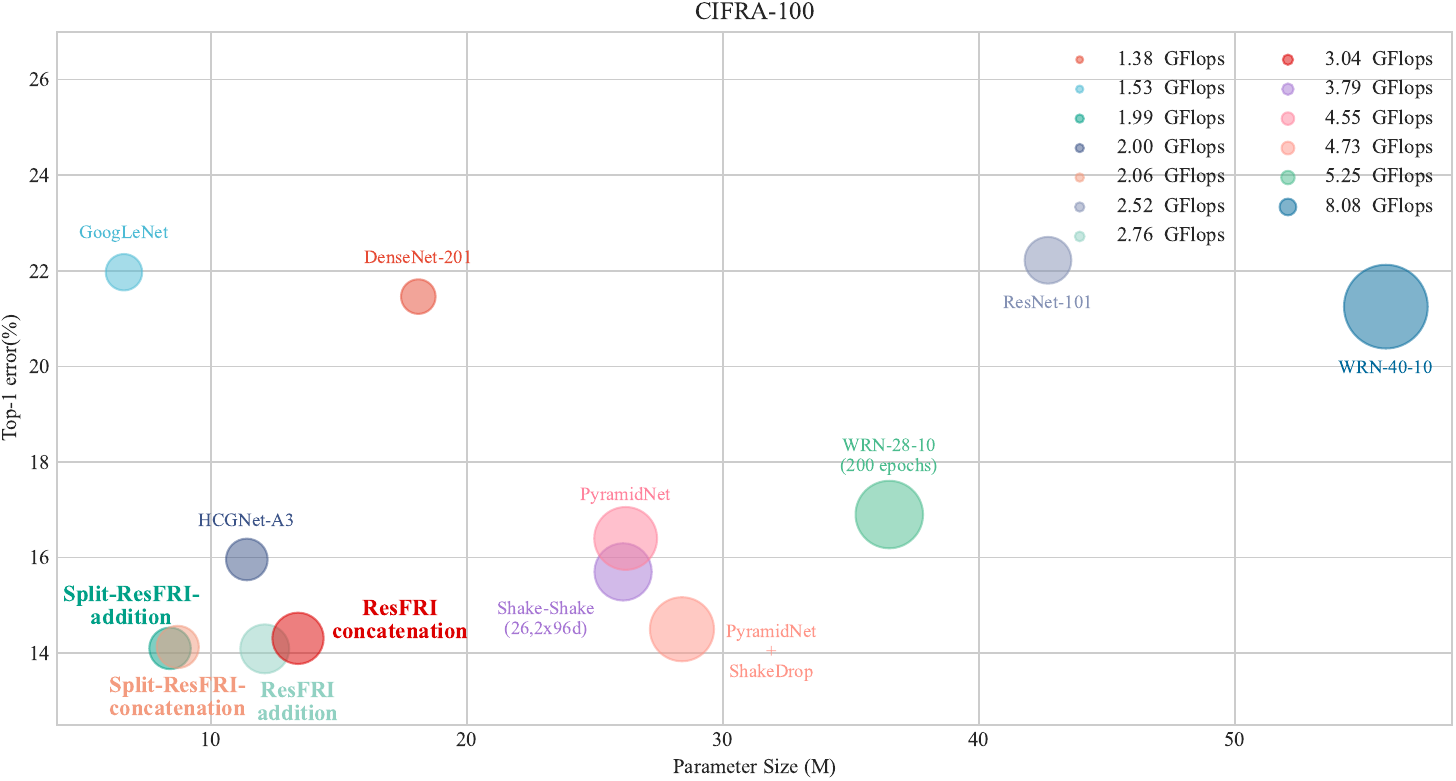}
	\caption{Comparisons of models on CIFAR100 Dataset }
	\label{CIFAR100}
\end{figure*}

The CIFAR100 dataset consists of $50$k training images and $10$k testing images from $100$ classes and their resolution is $32$$\times$$32$. And the detailed results of comparisons of different models will be clearly provided in Table \ref{table22} and Fig.\ref{CIFAR100}. 

\begin{table*}[htbp]\scriptsize
	\centering
	\caption{Top-1, Top-5 Test Error (\%) and Model Size on the CIFAR-100 Dataset}
	\renewcommand\arraystretch{1.2} %??????
	\begin{tabular}{ccccccccc} %????????c
		\hline 
		\multicolumn{4}{l}{Method}& Flops &  Params  &    top-1 err. &    top-5 err. & Accuracy\\ \cline{1-9}
		%		\multicolumn{4}{c}{Resnet18} &     11.2M    &      24.39        &6.95\\
		%		\multicolumn{4}{c}{Resnet34} &     21.3M    &      23.24        &6.63\\
		%		\multicolumn{4}{c}{Resnet50} &     23.7M    &      22.61        &6.04\\
		\multicolumn{4}{l}{ResNet-101 \cite{DBLP:conf/cvpr/HeZRS16}}&2.52G &    42.7M    &      22.22        &5.61 & 77.78\\
		%		\multicolumn{4}{c}{Resnet152} &    58.3M    &      22.31        &5.81\\
		%\multicolumn{4}{l}{ResNeXt-50 \cite{DBLP:conf/cvpr/XieGDTH17}}&- &     14.8M    &      22.23        &6.00\\
		%\multicolumn{4}{l}{ResNeXt-101 \cite{DBLP:conf/cvpr/XieGDTH17}}&- &     25.3M    &      22.22        &5.99\\
		%\multicolumn{4}{l}{ResNeXt-152 \cite{DBLP:conf/cvpr/XieGDTH17}} &  -&   33.3M    &      22.40        &5.58\\
		\multicolumn{4}{l}{DenseNet (k=12, depth=40) \cite{DBLP:journals/pami/HuangLPMW22}} & - &  1.0M    &      27.55       &-&72.45\\
		\multicolumn{4}{l}{DenseNet (k=12, depth=100)\cite{DBLP:journals/pami/HuangLPMW22}} &   -& 7.0M    &      23.79       &-&76.21\\
		\multicolumn{4}{l}{DenseNet (k=24, depth=100)\cite{DBLP:journals/pami/HuangLPMW22}}&-&    27.2M    &      23.42        &-&76.58\\
		\multicolumn{4}{l}{DenseNet-BC (k=24) \cite{DBLP:journals/pami/HuangLPMW22}}&- &    15.3M     &      19.64        &-&80.36\\
		\multicolumn{4}{l}{GoogLeNet \cite{DBLP:conf/cvpr/SzegedyLJSRAEVR15}}&1.53G &     6.6M    &      21.97        &5.94 & 78.03\\
		%\multicolumn{4}{l}{Inception v3 \cite{DBLP:conf/cvpr/SzegedyVISW16}}&- &     22.3M    &      22.81        &6.39\\
		\multicolumn{4}{l}{Inception v4 \cite{DBLP:conf/aaai/SzegedyIVA17}}&- &     41.3M    &      24.14        &6.90 & 75.86\\
		\multicolumn{4}{l}{InceptionResnet v2 \cite{DBLP:conf/aaai/SzegedyIVA17}}&- &     65.4M    &      27.51        &9.11 & 72.49 \\
		%\multicolumn{4}{l}{Xception \cite{DBLP:conf/cvpr/Chollet17}}&- &     21.0M    &      25.07        &7.32\\
		\multicolumn{4}{l}{LOW \cite{DBLP:journals/pr/SantiagoBSCN21}}&- &     -   &      22.8    &- & 77.2\\
		%\multicolumn{4}{l}{WRN-40-10 \cite{DBLP:conf/bmvc/ZagoruykoK16}}&8.08G &     55.9M    &      21.25        &5.77\\
		%		\cline{1-8}
		%\multicolumn{4}{l}{	FractalNet \cite{DBLP:conf/iclr/LarssonMS17}}&- &     38.6M   &      29.05 & -      \\
		\multicolumn{4}{l}{SOPCNN \cite{DBLP:conf/mldm/Assiri19}}&- &    4.2M    &      27.04        & - &72.96\\
		\multicolumn{4}{l}{FocusNet \cite{DBLP:journals/pr/ZhangSS22}}&4.46G&    11.3M     &      21.71     &- &78.29\\
		\multicolumn{4}{l}{AFI-ResNext-29(32 $\times$ 4d) \cite{DBLP:journals/nn/PanXPWLBFX22}}&21.74G&    4.26M     &      20.67     &- & 79.33\\
		\multicolumn{4}{l}{DCDN \cite{DBLP:journals/prl/PatelW22}}&- &     -   &      16.98      &-&83.02\\
		\multicolumn{4}{l}{WRN-28-10 \cite{DBLP:conf/icml/KwonKPC21}}&5.25G &     36.5M    &      16.9        &-&83.1\\
		%		\multicolumn{4}{l}{WRN-28-10 (200 epochs)}&5.25 GFlops &     36.5M    &      16.9        &-\\
		\multicolumn{4}{l}{Res2NeXt-29, 6c$\times$24w$\times$6s \cite{DBLP:journals/pami/GaoCZZYT21}}&- &     36.7M    &      16.79        &-&83.21\\
		\multicolumn{4}{l}{Res2NeXt-29, 6c$\times$24w$\times$6s-SE \cite{DBLP:journals/pami/GaoCZZYT21}}&- &     36.9M    &      16.56        &-&83.44\\
		\multicolumn{4}{l}{PyramidNet \cite{DBLP:conf/cvpr/HanKK17}}&4.55G &     26.2M    &      16.4       &-&83.6\\
		%\multicolumn{4}{l}{OR-WideResNet \cite{DBLP:conf/cvpr/ZhouYQJ17}}&- &     18.2M    &      16.15        &2.98\\
		\multicolumn{4}{l}{NASNet-A \cite{DBLP:conf/aaai/YangAZHZXLX20}}&- &     50.9M    &      16.03        &- &83.97\\
		\multicolumn{4}{l}{HCGNet-A3 \cite{DBLP:conf/aaai/YangAZHZXLX20}}&2.0G &     11.4M    &     15.96        &-&84.04\\
		\multicolumn{4}{l}{Shake-Shake (26 2$\times$96d) \cite{DBLP:conf/iclr/ForetKMN21}}&3.79G &     26.1M    &     15.7        &-&84.3\\
		\multicolumn{4}{l}{NAONet \cite{DBLP:journals/pr/HuBAYF22}}&- &    -  &     15.67      &-&84.33\\
		\multicolumn{4}{l}{PyramidNet+ShakeDrop \cite{DBLP:conf/iclr/ForetKMN21}}&4.73G &     28.4M    &     14.5     &-&85.5\\\cline{1-9}
		%		\multicolumn{4}{c}{EEEA-Net-C ($\beta = 5$) + CO \cite{DBLP:journals/eaai/TermritthikunJI21}} &     3.6M    &     15.02       &-\\
		%		\multicolumn{4}{l}{WRN-28-10 (200 epochs) + Cutout + Sam \cite{DBLP:conf/iclr/ForetKMN21}} &     36.5M    &      14.9        &-\\
		%		\multicolumn{4}{c}{Shake-Shake (26 2x96d)} &     -    &     14.30       &-\\\cline{1-7}
		
		\multicolumn{4}{l}{ResFRI-addition}&2.76G &    12.1M 	    &      \textbf{14.09}        &2.42&85.91\\
		\multicolumn{4}{l}{Split-ResFRI-addition}& 1.99G &      8.4M   &      14.10& 2.48&85.90      \\
		\multicolumn{4}{l}{ResFRI-concatenation}&3.04G &     13.4M     &      14.31 & 2.71 &85.69\\
		\multicolumn{4}{l}{Split-ResFRI-concatenation}&2.06G &     8.7M     &   14.13 & \textbf{2.32} & 85.87\\
		\hline
	\end{tabular}
	\label{table22}    
\end{table*}

By checking the results given in Table \ref{table22}, some conclusions can be made. ResNet-101 has a performance lag of around $8$\% compared with the proposed method while it utilizes approximate flops and nearly three times parameter amount of ResFRI. For Res2NeXt-series models, all of them achieves analogous performance as ResNet-101 \cite{DBLP:conf/cvpr/HeZRS16} with much less flops and parameter amounts. The situation of DenseNets \cite{DBLP:journals/pami/HuangLPMW22} is also similar, they further reduce the size and computational complexity of the model, but the accuracy of them is still at a comparatively low level. The best accuracy has at least a performance disadvantage of $5$\% compared with ResFRI-series models. Nevertheless, the inception-series models have relatively excellent performances. Particularly, GoogLeNet \cite{DBLP:conf/cvpr/SzegedyLJSRAEVR15} possesses only $6.6$M parameter amount but achieves an effect that is ahead of many models. Considering the results of WRN-28-10 provided in \cite{DBLP:conf/icml/KwonKPC21}, it achieves a performance leap with a top-1 error rate of about $16$\% and dose not increase flops and parameters amount too much compared with the previous models. And it can be obtained that Res2NeXt can reach a similar performance with roughly the same number of parameters as WRN-28-10. In general, although the performance of the models is acceptable,  ResFRI and Split-ResFRI have higher accuracy with much lower flops and parameter amounts compared with the two categories of models we just discussed. 

Besides, for PyramidNet \cite{DBLP:conf/cvpr/HanKK17}, Shake-Shake (26 2$\times$96d) \cite{DBLP:conf/iclr/ForetKMN21} and PyramidNet+ShakeDrop \cite{DBLP:conf/iclr/ForetKMN21}, ResFRI and Split-ResFRI achieve better performances while using less flops and parameter amount. The most light one, Split-ResFRI-addition can achieve almost the best performance with less than $9$M parameter amount and $2$ Gflops which are between a half and a third of the scales of the four models mentioned before. Especially, the combination of PyramidNet and ShakeDrop has the closest performance to the proposed method while possessing roughly $55$\% higher parameter amount and $110$\% more flops than the proposed models at least. Compared with the original PyramidNet, the combination of PyramidNet and ShakeDrop obtains a performance improvement of about $2$\%, which illustrates the possibility of follow-up work using this technology and the effectiveness of ShakeDrop. All in all, based on experimental results provided in Table \ref{table22}, it can be concluded that the proposed method possesses a better precision on classification task when compared with classical networks. Except for GoogLeNet and DenseNet, all of the other models have bigger parameter amount than the proposed model but could not reach the same level of accuracy, which demonstrates the efficiency and effectiveness of the proposed model. Although GoogLeNet and DenseNet own smaller model scale than ResFRI and Split-ResFRI, our proposed method has a huge advantage in accuracy. Concretely, the version of addition of ResFRI reaches a top-1 error rate $14.09$ and top-5 error rate $2.42$ on CIFAR-100 dataset. In the meantime, Split-ResFRI can achieve very similar performance with at most $37.3$\% and $34.5$\% reduction of parameter amount and flops. In one word, the comparisons prove the superiority of ResFRI and Split-ResFRI on classification tasks which can be regarded as a satisfying solution in choices among CNN architectures.

\subsubsection{Comparison with models using extra techniques and data}

\begin{table*}[htbp]\scriptsize
	\centering
	\caption{Error rate (\%) and Model Size on the CIFAR-100 Dataset}
	\renewcommand\arraystretch{1.4} %??????
	\begin{tabular}{cccccc} %????????c
		\hline 
		 Method &     top-1 err.&Accuracy & Method with extra data  &    top-1 err. &Accuracy\\ \hline
		 WRN-28-10 (200 epochs) + SAM & 16.5& 83.5 & ViT-H/14 + JFT & 5.45& 94.55\\
		 WRN-28-10 (1800 epochs) + SAM & 16.3 & 83.7 & ViT-L/16 + JFT & 6.10& 93.90 \\
		 Shake-Shake (26 2x96d) + SAM & 15.1  & 84.9 & ViT-L/16 + I21k &  6.75& 93.25 \\
		 PyramidNet + SAM & 14.6 & 85.4 & ResNet152x4 + BiT-L   & 6.49& 93.51 \\ 
		 PyramidNet + ShakeDrop + SAM & 13.3  & 86.7 & $\mu$2Net + ViT L/16  &5.6& 94.4\\
	   EffNet-L2 + SAM + Pre-Training$^{*}$&3.92& 96.08 & DINOv2 + ViT-g/14& 5.05& 94.95\\\hline
        The Proposed Method&top-1 err.&Accuracy & The Proposed Method& top-1 err.&Accuracy\\\hline
            ResFRI-addition & 14.09 & 85.91 & Split-ResFRI-addition & 14.10& 85.90 \\
            ResFRI-concatenation & 14.31 & 85.69 &Split-ResFRI-concatenation & 14.13& 85.87\\
            ResFRI-addition + SAM & 12.31 & 87.69 & Split-ResFRI-addition + SAM &12.27& 87.73\\
            ResFRI-concatenation + SAM &12.45 & 87.55 &Split-ResFRI-concatenation + SAM &12.24& 87.76 \\\hline
	\end{tabular}
	\label{}    
\end{table*}

Compared with the model with approximate parameters, our model is able to achieve better results without using additional training means, and can also achieve considerable performance improvement after taking advantage of SAM. In comparison with remaining models, the situation is similar to the experimental comparison on CIFAR10 dataset. In general, the number of classes to distinguish has increased significantly in CIFAR100 dataset, and a model that realize excellent performance on CIFAR10 may not perform well on CIFAR100. However, the proposed model can also achieve very excellent classification accuracy on CIFAR100 dataset.

\subsection{Experiments on Tiny Imagenet}
The Tiny Imagenet dataset consists of $100$k training images and $10$k testing images from $200$ classes and their resolution is $64$$\times$$64$. And the results of comparisons are given in Table \ref{table2}. 

\begin{table}[htbp]\footnotesize
	\centering
	\caption{Top-1 Test Error (\%) and Model Size on the Tiny Imagenet Dataset}
	\renewcommand\arraystretch{1} %??????
	\begin{tabular}{cccccccc} %????????c
		\hline 
		\multicolumn{4}{l}{Method}& Flops &  Params  &    top-1 err. & Accuracy \\ \cline{1-8}
		%		\multicolumn{4}{l}{MMA-CCT-7/3x2}&- &     -   &    35.59      \\
		\multicolumn{4}{l}{DenseNet \cite{DBLP:journals/prl/PatelW22}}&-&    -     &      40.00 & 60.00     \\
		\multicolumn{4}{l}{CS-KD \cite{DBLP:journals/pr/ZhangSS22}}&-&    -    &      39.56  & 60.44  \\
		\multicolumn{4}{l}{ResNet-110 \cite{DBLP:journals/prl/PatelW22}}&-&    -     &      37.44 & 62.56     \\
		\multicolumn{4}{l}{BYOT \cite{DBLP:journals/pr/ZhangSS22}}&-&    -    &      36.90 & 63.10 \\
		\multicolumn{4}{l}{FRSKD \cite{DBLP:journals/pr/ZhangSS22}}&-&    -    &      35.65 & 64.35   \\
		\multicolumn{4}{l}{FocusNet \cite{DBLP:journals/pr/ZhangSS22}}&4.46Gflops&    11.3M     &      35.51 & 64.49     \\
		\multicolumn{4}{l}{Wide-ResNet \cite{DBLP:journals/prl/PatelW22}}&-&    -     &      34.01 & 65.99     \\
		\multicolumn{4}{l}{ResNext \cite{DBLP:journals/prl/PatelW22}}&-&    -    &      31.77 & 68.23     \\
		\multicolumn{4}{l}{DCDN \cite{DBLP:journals/prl/PatelW22}}&- &     77.79M   &      29.72  & 70.28   \\
		%		\multicolumn{4}{l}{	\makecell[l]{ResNeXt-50\\ + AutoMix}}&1.53 GFlops &    22.9M   &   29.28       \\
		\cline{1-8}
            
		%		\multicolumn{4}{c}{EEEA-Net-C ($\beta = 5$) + CO \cite{DBLP:journals/eaai/TermritthikunJI21}} &     3.6M    &     15.02       &-\\
		%		\multicolumn{4}{l}{WRN-28-10 (200 epochs) + Cutout + Sam \cite{DBLP:conf/iclr/ForetKMN21}} &     36.5M    &      14.9        &-\\
		%		\multicolumn{4}{c}{Shake-Shake (26 2x96d)} &     -    &     14.30       &-\\\cline{1-7}
		
		\multicolumn{4}{l}{ResFRI-addition (pruning ratio $0.7$)}&3.13G &    12.4M 	    &    31.5  & 68.5      \\
		\multicolumn{4}{l}{ResFRI-addition (pruning ratio $0$)}&3.13G &    12.4M 	    &      29.60  & 70.40    \\
		\multicolumn{4}{l}{Split-ResFRI-addition}& 2.37G&   8.5M   &   31.93 & 68.07   \\
		\multicolumn{4}{l}{ResFRI-concatenation}&3.4G &     13.7M     &      \textbf{29.46} & \textbf{70.54} \\
		\multicolumn{4}{l}{Split-ResFRI-concatenation}&2.44G & 9.0M &      32.04 & 67.96\\
		\hline
	\end{tabular}
	\label{table2}    
\end{table}

The comparative results are acquired from \cite{DBLP:journals/pr/ZhangSS22} and \cite{DBLP:journals/prl/PatelW22}. Specifically, the experiments on the Tiny Imagenet show that the proposed method achieves a considerably satisfying classification accuracy. For some traditional models like ResNet, DenseNet, Wide-ResNet and ResNext, both ResFRI and Split-ResFRI exceed their performance. And with respect some newly proposed methods, FocusNet \cite{DBLP:journals/pr/ZhangSS22} and DCDN \cite{DBLP:journals/prl/PatelW22}, the proposed models still achieve better results with much lower Flops and smaller parameter amount, which demonstrates the superiority of this proposed model. In sum, ResFRI and Split-ResFRI not only outperform other comparative models on  relative low-resolution datasets, but also achieve excellent results on dataset with higher pixels.

\subsubsection{Comparison with models using extra techniques and data}
Due to the limited computing resources, we are unable to perform larger-scale model training and utilize more data. Compared with the models in the table, the size of the model is much larger than the proposed method. For instance, the total number of Flops and parameters of ViT-L/16 are 56 to 80 times and 22 to 35 times of those of the model proposed in this paper. However, we reduce the number of parameters while ensuring a certain classification effect, which makes the model proposed in this paper be better applied in the actual environment.
\begin{table*}[htbp]\footnotesize
	\centering
	\caption{Error rate (\%) and Model Size on the Tiny Imagenet Dataset}
	\renewcommand\arraystretch{1.45} %??????
	\setlength{\tabcolsep}{0.75mm}{\begin{tabular}{cccccccccc} %????????c
		\hline 
		 Method & Flops&Params &    top-1 err. & Accuracy& Ours & Flops&Params &    top-1 err.& Accuracy \\ \hline
		 ViT-L/16 &190.7G &304M&13.57 & 86.43 &\makecell{Original Best\\ Performance} &3.4G& 13.7M&29.46 & 70.54\\
		 DeiT-B/16-D &55.5G &196M&12.71 & 87.29 & \makecell{Original Worst\\ Performance} &2.44G&9.0M&32.04 & 67.96 \\
		 \makecell{DeiT-B/16-D \\+ Tent, 10 iter} &55.5G&196M&  11.80& 88.20 &  \makecell{ResFRI\\-addition + OCD}&3.13G&12.4M& 27.36 & 72.64 \\
		 \makecell{DeiT-B/16-D \\+ Tent, 30 iter} &55.5G&196M& 11.80 &88.20&\makecell{Split-ResFRI\\-addition + OCD}   &2.37G&8.5M & 28.83& 71.17 \\ 
		  \makecell{DeiT-B/16-D \\+ OCD}&55.5G&196M&  9.20 & 90.80 & \makecell{ResFRI-\\concatenation + OCD}  &3.4G&13.7M& 27.10 & 72.90\\
		\makecell{DeiT-B/16-D\\ + OCD + ensemble} &55.5G&196M&8.00&92.00&\makecell{Split-ResFRI\\-concatenation + OCD}&2.44G&9.0M&29.49 & 70.51\\\hline
	\end{tabular}}
	\label{}    
\end{table*}

\subsection{Experiments on MNIST}
The MNIST dataset contains $60$k training images and $10$k testing images from $10$ classes whose resolution is $28$$\times$$28$. And the detailed results of comparisons of different models will be clearly provided in Table \ref{table3}.

%\begin{figure}[h]
%	\centering
%	\includegraphics[width = 0.47\textwidth]{MNSIT.pdf}
%	\label{MNIST}
%	\caption{Loss and Accuracy of GoogLe2Net on MNIST Dataset }
%\end{figure}

\begin{table}[htbp]\footnotesize
	\centering
	\caption{Test Accuracy (\%) and Model Size on the MNIST Dataset}
	\renewcommand\arraystretch{1.2} %??????
	\begin{tabular}{ccccccc} %????????c
		\hline 
		\multicolumn{4}{l}{Method} &  Params  &    top-1 err & Accuracy. \\ \cline{1-7}
		%		\cline{1-6}
		%\multicolumn{4}{l}{STBP \cite{articleWu}} &     -     &      0.58     \\
		%\multicolumn{4}{l}{Converted SNN \cite{articleRueckauer}} &     -     &      0.56    \\
		\multicolumn{4}{l}{Spiking CapsNet \cite{DBLP:journals/isci/ZhaoLZWZ22}} &     -     &      0.83  & 99.17  \\
		%\multicolumn{4}{l}{	CapsNet \cite{DBLP:conf/nips/SabourFH17}} &     -   &      0.25    \\
		%\multicolumn{4}{l}{	DropConnect \cite{DBLP:conf/icml/WanZZLF13}} &     -   &      0.21       \\
		\multicolumn{4}{l}{	RMDL \cite{DBLP:conf/icisdm/KowsariHBMB18}} &     -   &      0.18  & 99.82     \\
		\multicolumn{4}{l}{	SOPCNN \cite{DBLP:conf/mldm/Assiri19}} &     1.4M    &      \textbf{0.17} & 99.83     \\
		\multicolumn{4}{l}{FocusNet \cite{DBLP:journals/pr/ZhangSS22}}&    11.3M     &      0.34   & 99.66  \\
		\cline{1-7}
		%		\multicolumn{4}{c}{Shake-Shake (26 2x96d)} &     -    &     14.30       &-\\\cline{1-7}
		\multicolumn{4}{l}{ResFRI-addition} &     12.1M    &      0.35  & 99.65 \\
		\multicolumn{4}{l}{Split-ResFRI-addition}&       8.4M   &      0.39 & 99.61    \\
		\multicolumn{4}{l}{ResFRI-concatenation} &     13.4M    &    \textbf{ 0.31} & \textbf{99.69 }      \\
		\multicolumn{4}{l}{Split-ResFRI-concatenation} &     8.7M &    0.35 & 99.65 \\
		\hline
	\end{tabular}
	\label{table3}    
\end{table}

MNIST is one of the most famous dataset in machine learning and it is not taken into consideration of verification experiments of modern models due to its simplicity. In order to illustrate the performance of proposed method on simple dataset, we conduct this experiment and compare the results produced by it with some light and correspondingly-designed models. For RMDL \cite{DBLP:conf/icisdm/KowsariHBMB18} and SOPCNN \cite{DBLP:conf/mldm/Assiri19}, they obtain better results than the proposed model. Nevertheless, all of these models lack the ability to capture more complex characteristics, which is fully demonstrated by the experimental results on CIFAR-10 and CIFAR-100. On both of the datasets, the proposed model has an obvious performance advantage. Besides, with respect to some modern models such as FocusNet \cite{DBLP:journals/pr/ZhangSS22}, the proposed model still achieves similar or better results. The experiment on MNIST datasets proves that the proposed model has good performance in dealing with simple image problems.

\subsection{Experiments on FashionMNIST}
The FashionMNIST dataset consists of $60$k training images and $10$k testing images from $10$ classes and their resolution is $28$$\times$$28$. And the results of comparisons are given in Table \ref{table4}. 

%\begin{figure}[h]
%	\centering
%	\includegraphics[width = 0.47\textwidth]{FashionMNIST.pdf}
%	\label{FashionMNIST}
%	\caption{Loss and Accuracy of GoogLe2Net on FashionMNIST Dataset }
%\end{figure}

\begin{table}[htbp]\footnotesize
	\centering
	\caption{Test Accuracy (\%) and Model Size on the FashionMNIST Dataset\\}
	\renewcommand\arraystretch{1.2} %??????
	\begin{tabular}{ccccccc} %????????c
		\hline 
		\multicolumn{4}{l}{Method} &  Params  &    top-1 err. & Accuracy \\ \cline{1-7}
		\multicolumn{4}{l}{PreAct-ResNet18 \cite{DBLP:conf/eccv/HeZRS16}} &     11.1M     &      4.30  & 95.70    \\
		%		\multicolumn{4}{l}{WideResNet-28-10 + RandomErasing } &     37M     &      4.16       \\
		\multicolumn{4}{l}{WideResNet-28-10 \cite{DBLP:conf/icml/NoklandE19}} &     37M     &      4.16  & 95.84     \\
		\multicolumn{4}{l}{DenseNet-BC-190 \cite{DBLP:journals/pami/HuangLPMW22}} &     25.6M     &      4.06  & 95.94    \\
		%\multicolumn{4}{l}{STBP \cite{articleWu}} &     -     &      7.33      \\
		%\multicolumn{4}{l}{Converted SNN \cite{articleRueckauer}} &     -     &      7.33      \\
		\multicolumn{4}{l}{Spiking CapsNet \cite{DBLP:journals/isci/ZhaoLZWZ22}} &     -     &      8.93  & 91.07    \\
		\cline{1-7}
		%		\multicolumn{4}{c}{Shake-Shake (26 2x96d)} &     -    &     14.30       &-\\\cline{1-7}
		\multicolumn{4}{l}{ResFRI-addition} &     12.1M    &      4.00   & 96.00     \\
		\multicolumn{4}{l}{Split-ResFRI-addition}&       8.4M   &      \textbf{3.80} & \textbf{96.20}    \\
		\multicolumn{4}{l}{ResFRI-concatenation} &     13.4M    &     4.29 & 95.71       \\
		\multicolumn{4}{l}{Split-ResFRI-concatenation} &     8.7M     &   3.87 & 96.13 \\
		\hline
	\end{tabular}
	\label{table4}    
\end{table}
FashionMNIST serves as a more complex version of MNIST which is an important data set to measure the basic capacity of models. The proposed model outperforms traditional models such as PreAct-ResNet18 \cite{DBLP:conf/eccv/HeZRS16}, WideResNet-28-10 \cite{DBLP:conf/icml/NoklandE19} and DenseNet-BC-190 \cite{DBLP:journals/pami/HuangLPMW22}, which are with much lower parameter amounts. Besides, with respect to newer model like Spiking CapsNet \cite{DBLP:journals/isci/ZhaoLZWZ22}, the proposed model still achieves advantages by a large margin. In general, the proposed model achieves performance lead on this dataset and is a considerably light solution compared with other models.
\subsection{Experiments on SVHN}
The SVHN dataset contains $73257$ training images and $26032$ testing images from $10$ classes whose resolution is $32$$\times$$32$. And the detailed results of comparisons of different models will be clearly provided in Table \ref{table5}.

%\begin{figure}[h]
%	\centering
%	\includegraphics[width = 0.47\textwidth]{SVHN.pdf}
%	\label{SVHN}
%	\caption{Loss and Accuracy of GoogLe2Net on SVHN Dataset }
%\end{figure}

\begin{table}[htbp]\footnotesize
	\centering
	\caption{Test Accuracy (\%) and Model Size on the SVHN Dataset}
	\renewcommand\arraystretch{1.2} %??????
	\begin{tabular}{ccccccc} %????????c
		\hline 
		\multicolumn{4}{l}{Method} &  Params  &    top-1 err.&Accuracy \\ \cline{1-7}
		\multicolumn{4}{l}{ResNet-110 \cite{DBLP:journals/prl/PatelW22}}&    -     &      2.28  & 97.72    \\
		%\multicolumn{4}{l}{FractalNet \cite{DBLP:conf/iclr/LarssonMS17}} &     38.6M    &      2.01      \\
		%\multicolumn{4}{l}{FractalNet with Dropout/Drop-path \cite{DBLP:conf/iclr/LarssonMS17}} &     38.6M    &      1.87      \\
		\multicolumn{4}{l}{WRN-28-10 \cite{DBLP:journals/prl/PatelW22}}&    -     &      1.80   & 98.20   \\
		%		\multicolumn{4}{l}{Recurrent CNN} &     -    &      1.77      \\
		\multicolumn{4}{l}{ResNet with Stochastic Depth \cite{DBLP:conf/eccv/HuangSLSW16}} &     -   &      1.75 & 98.25     \\
		\multicolumn{4}{l}{DenseNet-BC \cite{DBLP:journals/pami/HuangLPMW22}} &     15.3M    &      1.74  & 98.26   \\\cline{1-7}
		%		\multicolumn{4}{c}{Shake-Shake (26 2x96d)} &     -    &     14.30       &-\\\cline{1-7}
		
		\multicolumn{4}{l}{ResFRI-addition} &     12.1M    &      \textbf{1.72}   & \textbf{98.28}     \\
		\multicolumn{4}{l}{Split-ResFRI-addition} &      8.4M   &      1.84 & 98.16    \\
		\multicolumn{4}{l}{ResFRI-concatenation} &     13.4M    &      1.75 & 98.25     \\
		\multicolumn{4}{l}{Split-ResFRI-concatenation} &     8.7M     &   1.82 & 98.18 \\
		\hline
	\end{tabular}
	\label{table5}    
\end{table}

SVHN dataset is derived from Google Street View Door Number. By analyzing the experimental results on SVHN dataset, the proposed model achieves relatively satisfying accuracy compared with other classic models. Besides, ResFRI utilizes $3$M less parameter amount than DenseNet-BC \cite{DBLP:journals/pami/HuangLPMW22} achieving better classification results on FashinMNIST dataset. Similar to situations on other datasets, ResFRI and Split-ResFRI also obtain satisfying accuracy level with relatively lower parameter amount on SVHN dataset. In all, the proposed model can be regarded as a more cost-effective solution in handling image-related problems.

\begin{table}[htbp]\scriptsize
	\centering
	\caption{Comparison Among ResFRI Variants' Performance on CIFAR10 Dataset}
	\renewcommand\arraystretch{1.5} %??\begin{adjustbox}{angle=90}
		\setlength{\tabcolsep}{1mm}{\begin{tabular}{cccccccccccccc} %????????c
			\hline 
			\multicolumn{4}{c}{Variants} &  Params  &    top-1 err.& Accuracy&\multicolumn{4}{c}{Variants} &  Params  &    top-1 err. & Accuracy\\ \hline
			\multicolumn{4}{c}{\makecell{ResFRI\\ (addition, pruning ratio 0.7)}}&    12.1M    &     2.08 & 97.92&\multicolumn{4}{c}{\makecell{ResFRI without AvgPooling layer\\ (addition, pruning ratio 0.7)}}  &   12.1M    &      2.25 & 97.75     \\
			\multicolumn{4}{c}{\makecell{ResFRI\\ (addition, pruning ratio 0.35)}} &    12.1M    &      2.29& 97.71 &\multicolumn{4}{c}{\makecell{ResFRI without residual connection\\ (addition, pruning ratio 0.7)}}  &8.9M    &      2.43 & 97.57  \\
			\multicolumn{4}{c}{\makecell{ResFRI\\ (addition, pruning ratio 0)}}&    12.1M    &      2.13 & 97.87  & \multicolumn{4}{c}{\makecell{ResFRI without transverse passages\\ (addition, pruning ratio 0.7)}}  &9.4M   &      2.12 & 97.88   \\
			\multicolumn{4}{c}{\makecell{ResFRI \\(concatenation, pruning ratio 0.7)}}&     13.4M     &      2.14& 97.86&\multicolumn{4}{c}{\makecell{ResFRI without AvgPooling layer\\ (concatenation, pruning ratio $0$))}} &     13.4M    &      2.30 & 97.70 \\
			\multicolumn{4}{c}{\makecell{ResFRI \\(concatenation, pruning ratio 0.35)}} &     13.4M     &      2.23 & 97.77 & \multicolumn{4}{c}{\makecell{ResFRI without residual connection\\ (concatenation, pruning ratio $0$)}}&     10.2M    &      2.37 & 97.63 \\
			\multicolumn{4}{c}{\makecell{ResFRI\\ (concatenation, pruning ratio 0)}} &     13.4M     &      \textbf{2.06}&\textbf{97.94}&\multicolumn{4}{c}{\makecell{ResFRI without transverse passages \\(concatenation, pruning ratio $0$)}} &     9.4M   &      2.60 & 97.40 \\\hline
		\end{tabular}}
%	\end{adjustbox}
	\label{table6}    
\end{table}

\subsection{Ablation Experiment}
In this section, we conduct the ablation experiment from two main aspects which are addition and concatenation version of ResFRI. In the preliminary stage of our experiment, we notice that for the addition version of ResFRI, a proper ratio of pruning may help to promote the accuracy of the model. And in the version of concatenation, no pruning may further enhance performance of the network. Therefore, all of the ablation experiments are not only involved with adjustment of structure of networks, but also the ratios of pruning. And all of the results are provided in the following Table \ref{table6}.

In detail, we remove three key components of ResFRI, namely AvgPooling layer, Residual connection and passages between groups of convolutional layers respectively, which are to verify their influence on performance of the proposed network. And we can find that when each of them is removed, the performance will encounter a precision loss to some extent. It strongly demonstrates that when all of those components are synthesized, the lowest top1-error can be reached. Moreover, we also investigate influences of different pruning ratio on precision of the proposed model, which also proves the rationality of our settings of ResFRI. 

\subsection{Implication of the proposed method in real-world applications}
Image classification models are widely used in the real world and they can help us solve many complex problems. For example, in the field of driverless cars, image classification models enables the car recognize objects on the road, including vehicles, pedestrians, road markings, etc., so that the car is able to safely navigate and avoid accidents. In the medical field \cite{he2024matrixbaseddistancepythagoreanfuzzy}, image classification models can help doctors automatically diagnose diseases, such as skin cancer, thereby improving the accuracy and efficiency of diagnosis. Moreover, with respect to the field of security, image classification models can help us automatically identify and monitor abnormal behavior and suspicious people in the crowd, so that safety and response speed can be improved. In addition, image classification models can help us with traffic monitoring and congestion detection, as well as commercial applications such as customer segmentation and product recommendation. Except for the application areas mentioned above, there are many other practical applications of image classification models. For instance, in the field of environmental monitoring, image classification models can help us detect environmental changes such as deforestation and urbanization, thus providing data support for environmental protection \cite{he2021conflicting, li2024high-fidelity, he2023tdqmf}. And in the field of scientific research, image classification models are helpful for us to automatically analyze and classify image data in astronomy, biology, geology and other fields, therefore accelerating the progress of scientific research. Besides, the image classification model can also be applied to image retrieval, intelligent transportation, smart home and many other domains, bringing convenience and intelligence to people's lives. With the continuous development of technology, the application prospect of image classification model will be more and more broad \cite{he2024generalized, li2022nndf, jiang2024mfdnet, he2022mmget, he2022ordinal}. In the future, we can expect to see more innovative applications, such as the use of image classification models for intelligence and automation in areas such as smart manufacturing, smart city planning, and smart security. At the same time, with the continuous improvement of data quality and algorithm model, the accuracy and reliability of image classification model will be further improved, which provide better supports and guarantees for practical applications. As a result, we believe that the ResFRI and Split-ResFRI modules proposed in this paper which achieve very excellent image classification performance can be widely applied in fields such as shadow removal \cite{li2023high-resolution} and decision making \cite{he2022new, he2023ordinal} to help people solve related problems more intelligently. And the usage of effective optimization algorithms may be helpful to better adapt the model to practival situations \cite{zare2023global,abualigah2023modified,xu2023spatio, agushaka2022dwarf}. More importantly, the model proposed in this paper has the characteristics of less parameters and computation resources required, which enables it to be applied to more complex and demanding scenarios and better meet different types of demands.

\section{CONCLUSIONS}
The efficiency of multi-scale design in CNNs is repeatedly verified by lots of modern models, especially the structure of incpetion-like networks. However, the potentials of inception-based models are far from the upper bound. To effectively reuse feature with different receptive fields inside inception block, ResFRI and Split-ResFRI are proposed in which the passages between adjacent groups of convolutional layers are designed to realize the purpose and residual connection may help further improve performance of the network compared with original inception framework. All in all, ResFRI and Split-ResFRI are able to achieve better image classification accuracy with fewer parameters and flops and may be considered as an effective backbone in lots of other aspects of computer vision such as semantic segmentation and object detection in future works. In addition, researchers can easily integrate ResFRI and Split-ResFRI into other inception-like networks without any migration cost and may reduce overall complexity of the whole model benefiting from the lightweight design of the proposed model. We believe the combination of concept of multi-scale and residual connection in CNN architecture will further boost model performance on various vision tasks and be a research hot spot in the field of computer vision.

For future work, we plan to improve our work from two aspects. For the first, the way of information interaction between different processing branches is an open issue. Except for simple concatenation and addition of feature map, a more efficient policy of fusion of information can be devised which is one of the research focuses of our follow-up work. Moreover, in this paper, we investigate the influence brought by split of the information in channels whose optimal ratio is not completely clear, which may lead to fluctuation in model performance. It is worth noting that a proper split ratio may bring improvement of accuracy and reduce demand of calculation resources of models. As a result, we will also explore the relationship between model performance and information split ratio to optimize structure of networks. Synthesizing improvements on the two aspects, a more efficient and effective network architecture can be designed.

%\subsubsection*{Author Contributions}
%If you'd like to, you may include  a section for author contributions as is done
%in many journals. This is optional and at the discretion of the authors.

\subsubsection*{Acknowledgments}
This work is supported by National Science and Technology Major Project (Granted No. 2020AAA0109401) and Natural Science Foundation of China (granted No. 62192731). And the authors greatly appreciate the anonymous reviewers' suggestions and the editor's encouragement.
\bibliographystyle{elsarticle-num}
\bibliography{cite}
\end{document}